# Improving digital signal interpolation: $L_2$-optimal kernels with kernel-invariant interpolation speed

Oleg S. Pianykh, PhD

**Abstract**

Interpolation is responsible for digital signal resampling and can significantly degrade the original signal quality if not done properly. For many years, optimal interpolation algorithms were sought within constrained classes of interpolation kernel functions. We derive a new family of unconstrained $L_2$-optimal interpolation kernels, and compare their properties to the previously known.

Although digital images are used to illustrate this work, our $L_2$-optimal kernels can be applied to interpolate any digital signals.

*Key words*: interpolation, Fourier transform, $L_2$ space.

## 1 Introduction

Digital signals are acquired at discrete sampling points. For example, two-dimensional digital images are defined as fixed-size pixel matrices $I(k,n)$, where integer indices $k$ and $n$ (sampling coordinates) span over horizontal and vertical image axes, and the $I(k,n)$ value reflects image intensity of a given pixel $(k,n)$. As a result, rotating or zooming $I(k,n)$ implies recalculating its pixel values on a different (rotated or zoomed) pixel grid—that is, interpolating image pixel values at arbitrary $(x,y)$ coordinates.

The standard approach to this pixel resampling interprets image features as Fourier frequencies and interpolation as frequency-preserving filtering [2], [11], [12]. This means that the interpolated image $J(x,y)$ is computed as the convolution of the original pixels $I(x,y)$ with continuous 2D interpolation kernel function $h_{2D}(x,y)$:

$$J(x,y) = \sum_k \sum_n I(k,n) \times h_{2D}(x-k, y-n) \qquad \text{(Eq. 1)}$$

The most basic requirements traditionally imposed on $h_{2D}(x,y)$ are separability, symmetry, and finite support:

$$\begin{aligned} h_{2D}(x,y) &= h(x) \times h(y), \\ h(x) &= h(-x), \\ h(x) &= 0 \quad \forall x : |x| > L, \quad 0 < L < \infty \end{aligned} \qquad \text{(Eq. 2)}$$

Kernel symmetry implies filter isotropy, finite support $L$ – computability,[1] and separability reduces (Eq. 1) to a shorter one-dimensional convolution:

$$J(x,y) = \sum_{k=0}^{L} h(x-k) \left[ \sum_{n=0}^{L} I(k,n) \times h(y-n) \right] \qquad \text{(Eq. 3)}$$

In addition to this, we want (Eq. 3) to preserve the $I(x,y)$ average and remain identity for the overlapping pixel values (consider $(x,y)=(k,n)$), which leads to the following fundamental interpolating kernel conditions [2]:

$$h(n) = \delta_n^0,$$
$$\sum_{k=-L}^{L} h(x+k) = 1, \quad 0 \leq x \leq 1, \quad n,k \in Z, \quad L \in N \qquad \text{(Eq. 4)}$$

Conditions in (Eq. 2) and (Eq. 4) therefore are considered required for any interpolation kernel $h(x)$ [2], [3]. However, they can be satisfied by vast classes of functions. As a result, interpolation theory has been driven by the study of additional optimality constraints, leading to unique determination of $h(x)$: kernel smoothness, order of approximation, kernel size, and computational complexity (to name a few). This work produced a cohort of widely accepted kernels, summarized in [2]–[4]. With few exceptions, all those $h(x)$ values were sought in the space of simple-to-compute piecewise polynomials [1], leading to such well-known interpolation kernels as Keys, B-splines, or MOMS[2] [2]–[8]. In digital imaging applications, the following kernels[3] have become particularly popular:

$$\text{Linear:} \quad h_{Linear}(x) = 1 - x, \quad 0 \leq x \leq 1, \quad L = 1$$

$$\text{Keys:} \quad h_{Keys}(x) = \begin{cases} (a+2)x^3 - (a+3)x^2 + 1, & 0 \leq x \leq 1 \\ ax^3 - 5ax^2 + 8ax - 4a, & 1 \leq x \leq 2 \end{cases} \quad a = -\tfrac{1}{2}, \quad L = 2 \qquad \text{(Eq. 5)}$$

$$\text{Cubic3:} \quad h_{Cubic3}(x) = \frac{1}{5} \begin{cases} 6x^3 - 11x^2 + 5, & 0 \leq x \leq 1 \\ -3x^3 + 16x^2 - 27x + 14, & 1 \leq x \leq 2 \\ x^3 - 8x^2 + 21x - 18, & 2 \leq x \leq 3 \end{cases} \quad L = 3$$

These choices of $h(x)$ ruled digital image interpolation for decades [1], [7], [11], mainly due to their computational simplicity. Nevertheless, the evident advances in processor power made most of the early kernel criteria outdated, thus inviting a less constrained analysis of interpolation kernel optimality. With this in mind, we derive completely unconstrained $L_2$-optimal interpolation kernels, and compare their numerical properties to the interpolation classics in (Eq. 5).

---

[1] In most applications, support $L$ is commonly chosen as 1, 2, or 3 and rarely 4. $L=2$ and $L=3$ are the most typical choices in imaging software; DirectX and OpenGL still rely on $L=1$.

[2] MOMS kernels relax interpolation conditions (Eq. 4) to provide optimal order of interpolation.

[3] For notational simplicity we provide only non-zero segments of $h(x)$.

## 2 L$_2$-optimal interpolation kernels

Fourier transform of the *Sinc* kernel $h_s(x) = \sin(\pi x)/(\pi x) = Sinc(x)$ produces the ideal rectangular ("box") frequency response:

$$\Pi(t) = \begin{cases} 1, & t < 1/2 \\ 1/2, & t = 1/2 \\ 0, & t > 1/2 \end{cases}$$

In this respect, *Sinc(x)* acts as the most frequency-preserving interpolation kernel satisfying (Eq. 4), but unfortunately has infinite support *L*, and cannot be used practically. Therefore, previous interpolation kernel research concentrated on building non-*Sinc* kernels *h(x)* with other optimal properties (such as smoothness, order of approximation, and proximity to *L*-truncated *Sinc*) achievable on finite support. This approach has proven to be very fruitful, generating a wealth of kernel functions, designs, and optimality criteria [2].

We remove any additional constraints to find the most frequency-preserving *h(x)*, satisfying only the basic kernel conditions in (Eq. 2) and (Eq. 4).

To do so, we introduce *L$_2$-optimal interpolation kernels* as a theoretically-optimal way of preserving image frequency content with interpolation in (Eq. 3). For a positive integer *L*, let $\Lambda_L$ be the set of all symmetric *h(x)* with finite support *L*, satisfying (Eq. 4). For $\forall h(x) \in \Lambda_L$ and its Fourier transform

$$F_h(t) = \int_{-\infty}^{\infty} h(x) e^{-2\pi i x t} dx$$

we define the *frequency approximation error* (FAE) function as

$$E(h) = \left( \int_{t=-\infty}^{+\infty} (F_h(t) - \Pi(t))^2 dt \right)^{1/2}$$

We define L$_2$-*optimal interpolation kernel* $H_L(x) \in \Lambda_L$ as

$$H_L(x) = \arg\min_{h(x) \in \Lambda_L} E(h) \qquad \text{(Eq. 6)}$$

FAE function *E(h)* measures the accuracy of *h(x)* interpolation in the frequency domain. By definition *E(Sinc)* = 0 (ideal interpolation with no frequency loss), and *E(0)* = 1 (*h(x)* = 0 ("no interpolation" case, full data loss). Note that we can also study FAE as function of *L*: let us define $E_L(L) = E(H_L)$ – minimal FAE for given support size *L*. Then in terms of the kernel support size we rephrase *h(x)* = *Sinc(x)* and *h(x)* = 0 cases as $E_L(\infty) = 0$ and $E_L(0) = 1$. We expect $E_L(L)$ to decay with *L* (larger support means better interpolation), but we would like to find how exactly $E_L(L)$ depends on *L*.

### Theorem 1 (L$_2$-optimal interpolation):
*For any support size L, the optimal L$_2$-kernel H$_L$(x) in (Eq. 6) exists and is uniquely defined by the following continuous function:*

$$H_L(x) = \begin{cases} Sinc(x) + \dfrac{1}{2L} \left[ 1 - \sum_{k=0}^{2L-1} Sinc\left( (-1)^k x + \left\lfloor \dfrac{k+1}{2} \right\rfloor - (-1)^k n \right) \right] \\ n \leq x < n+1, \quad n = 0 \ldots L-1 \end{cases} \qquad \text{(Eq. 7)}$$

The full proof of Theorem 1 can be found in Appendix. As one can see, on each segment $[n, n+1]$, optimal $H_L(x)$ consists of the "ideal kernel" $Sinc(x)$ and the "aliasing" term $T_n(x)$ (penalizing $H_L(x)$ for the finite support $L$):

$$T_L(x) = H_L(x) - Sinc(x) = \begin{cases} T_n(x), \\ n \leq x < n+1, \; n = 0...L-1 \end{cases} = \begin{cases} \frac{1}{2L}\left[1 - \sum_{k=0}^{2L-1} Sinc\left((-1)^k x + \left\lfloor\frac{k+1}{2}\right\rfloor - (-1)^k n\right)\right] \\ n \leq x < n+1, \; n = 0...L-1 \end{cases} \quad (Eq.\ 8)$$

$T_L(x)$ defines, how much $L$-supported $H_L(x)$ differs from the ideal $Sinc(x)$, therefore $T_L(x)$ vanishes to 0 as $L$ increases (see Figure 1, left).

Knowing the exact formula for the least-squares-optimal $H_L(x)$ enables us to derive more general conclusions about the interpolation kernel optimality. In particular, we can find $E_L(L)$: the minimal frequency approximation error as a function of the kernel support size $L$.

**Corollary 1:**
*Minimal frequency approximation error (FAE) function for finite support L is given by (Figure 1, right):*

$$E_L = E(H_L) = \sqrt{2\int_0^L T_L^2(x)dx + 2\int_L^{+\infty} Sinc^2(x)dx} = \sqrt{2\sum_{n=0}^{L-1}\left[\int_n^{n+1} T_n^2(x)dx\right] + 2\int_L^{+\infty} Sinc^2(x)dx} \quad (Eq.\ 9)$$

Proof

By substituting the optimal $h(x)=H_L(x)$ from (Eq. 7) into (Eq. 18). Note that "no interpolation" case for $L=0$ is also included: $E_L(0) = \sqrt{2\int_0^{+\infty} Sinc^2(x)dx} = \sqrt{2\int_0^{+\infty} \Pi^2(t)dt} = 1$, as we described previously.

∎

The practical meaning of FAE $E_L=E(H_L(x))$ is straightforward: any $L$-supported interpolation kernel $h(x)$ can reproduce at most $1-E_L$ of the original image frequency content (measured in $L_2$ norm). Therefore, we believe that $E_L(L)$ gives better measure of kernel quality than smoothness or boundary conditions (such as maximum order)—it simply reflects how close we get to the ideal $\Pi(t)$ filter in the least-squares sense.

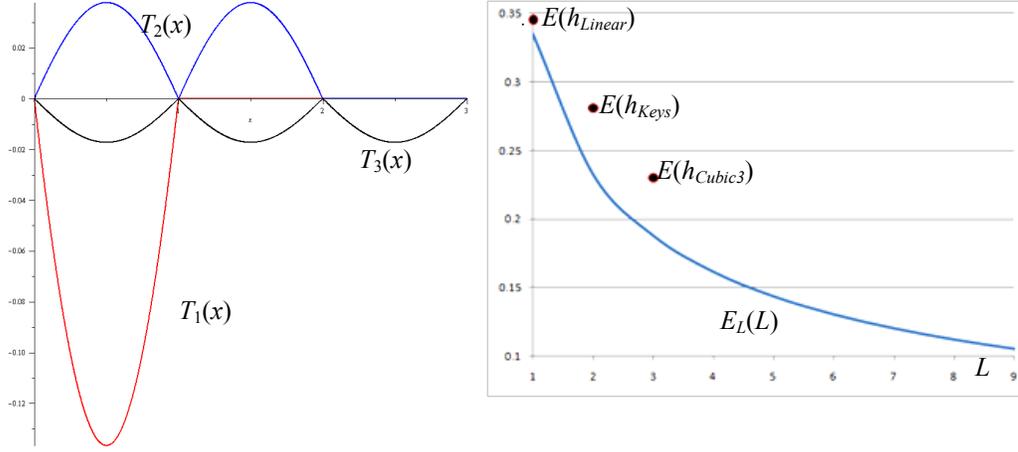

Figure 1: Left: $T_L(x)=H_L(x)-Sinc(x)$ for $L=1$ (red), $L=2$ (blue), and $L=3$ (black). Right: Optimal error $E_L(L)$ for $L = 1\ldots 9$, and its suboptimal values for the kernels in (Eq. 5).

For practical use, we have found the following approximate formula for (Eq. 9):

$$\hat{E}_L = 0.33 L^{-0.5258} \qquad \text{(Eq. 10)}$$

which for $L \leq 15$ deviates from the true (Eq. 9) by less than 2%.

## 3 Kernel comparison

### 3.1 Analysis

As you can see in Figure 1(right), the steepest reduction in $E_L(L)$ occurs for small $L$. Moreover, we are most interested in small $L$ to minimize the computational effort in (Eq. 3). Therefore, we chose to compare our optimal kernels in (Eq. 7) with the popular ones in (Eq. 5) for L = 1, 2, and 3. From (Eq. 7) we derive:

**Corollary 2:**
$L_2$-optimal interpolating kernels $H_1(x)$, $H_2(x)$, and $H_3(x)$ (for support L=1, 2, and 3 respectively) are

$$H_1(x) = \tfrac{1}{2}(1 + Sinc(x) - Sinc(1-x)), \quad 0 \leq x \leq 1$$

$$H_2(x) = \tfrac{1}{4}\begin{cases} 1 + 3Sinc(x) - Sinc(1-x) - Sinc(1+x) - Sinc(2-x), & 0 \leq x \leq 1 \\ 1 + 3Sinc(x) - Sinc(1-x) - Sinc(2-x) - Sinc(3-x), & 1 \leq x \leq 2 \end{cases} \qquad \text{(Eq. 11)}$$

$$H_3(x) = \tfrac{1}{6}\begin{cases} 1 + 5Sinc(x) - Sinc(1-x) - Sinc(1+x) - Sinc(2-x) - Sinc(2+x) - Sinc(3-x), & 0 \leq x \leq 1 \\ 1 + 5Sinc(x) - Sinc(1-x) - Sinc(1+x) - Sinc(2-x) - Sinc(3-x) - Sinc(4-x), & 1 \leq x \leq 2 \\ 1 + 5Sinc(x) - Sinc(1-x) - Sinc(2-x) - Sinc(3-x) - Sinc(4-x) - Sinc(5-x), & 2 \leq x \leq 3 \end{cases}$$

These kernels are shown in Figure 2 (left), along with their Fourier transforms (right). As expected, all $H_k(x)$ satisfy (Eq. 4), and their Fourier transforms provide the best least-squares approximation to $\Pi(t)$ for the given support size $L$.

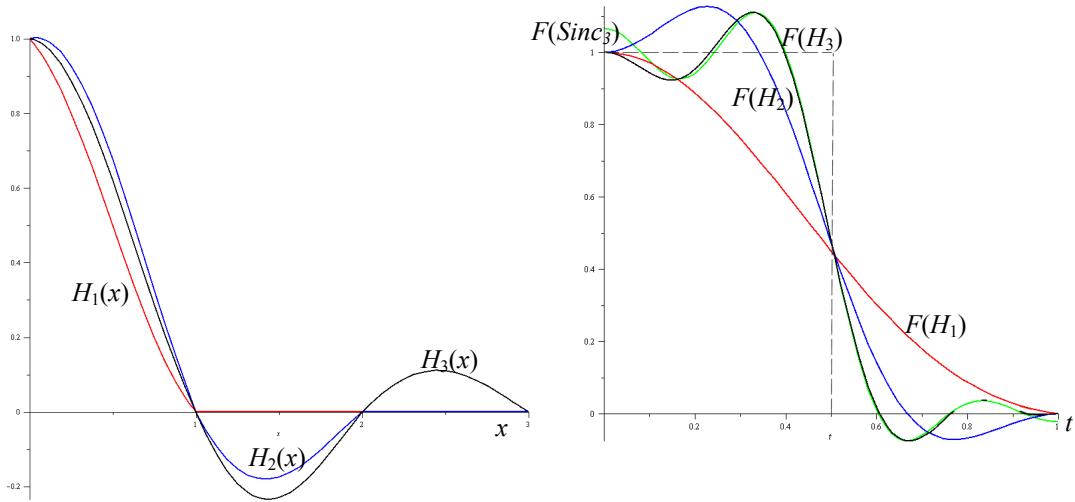

Figure 2: Left: $L_2$-optimal kernels $H_1(x)$ (red), $H_2(x)$ (blue), and $H_3(x)$ (black) Right: Their Fourier transforms. Fourier transform graph also includes the Fourier transform for $Sinc_3(x)$ (green) - $Sinc(x)$, truncated to finite $L=3$ support.

Note that $H_1(x)$ is very close to the linear interpolation kernel $h_{Linear}(x) = 1-x$, and $H_3(x)$ would be undistinguishable from $Sinc(x)$ on the left Figure 2 plot (which is why we do not show $Sinc(x)$ there). However, their differences become more apparent in the frequency domain—compare $F(H_3)$ and $F(Sinc_3)$ for small $t$ in Figure 2 right. For this reason, we compare the optimal $H_k(x)$ and popular (Eq. 5) kernels in the Fourier domain, as illustrated in Figure 3.

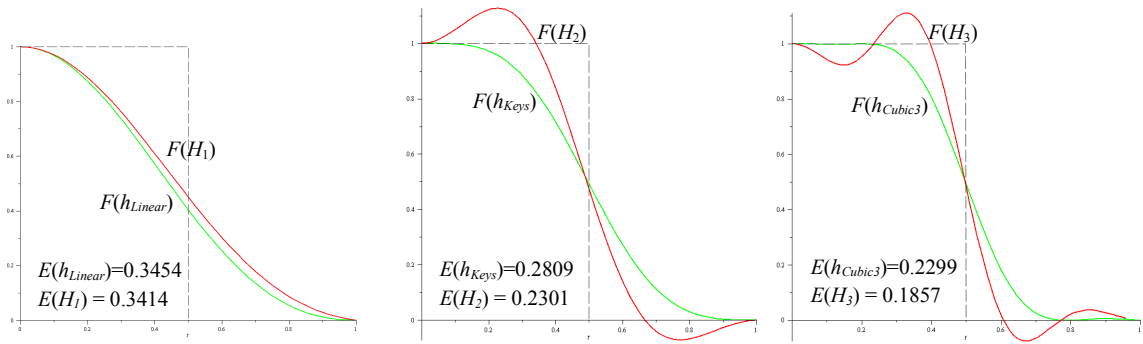

Figure 3: Comparing $H_k(x)$ to popular interpolation kernels in Fourier domain: $H_1(x)$ vs. $h_{Linear}(x)$ ($L=1$), $H_2(x)$ vs. $h_{Keys}(x)$ ($L=2$), $H_3(x)$ vs. $h_{Cubic3}(x)$ ($L=3$). Fourier transforms for $H_k(x)$ are shown in red, and FAE values $E(h)$ are shown for each kernel $h$.

In particular, all Fourier transforms $F(H_k)$ for $k > 1$ have local extrema, while the traditional kernel design preference was to avoid them, forcing (nearly) monotone $F(h)$. However, we believe that Fourier monotonicity is a rather subjective choice: it will not make the $F(h)$ frequencies more equal unless $F(h)$ is as flat as $\Pi(t)$. Therefore, while substantial peaks in $F(h)$ should be avoided,[4] balancing them around constant $\Pi(t)$ segments can make more practical sense.

---

[4] Large peaks in $F(h)$ are also responsible for interpolation "ringing" artifacts.

Another interesting observation can be derived from comparing the frequency approximation errors $E(h)$ for optimal (Eq. 11) and popular (Eq. 5) kernels ($E$-values in Figure 3). Not only did the optimal kernels produce smaller errors (as expected), but in some cases optimal kernels on smaller support ($H_2(x)$ for $L = 2$, $E(H_2) = 0.2301$) can preserve more frequency content than the best-known kernels on larger support ($h_{Cubic3}(x)$ on $L = 3$, $E(h_{Cubic3}) = 0.2299$). This strongly speaks in favor of the optimal kernels; using smaller support sizes, they deliver faster interpolation while preserving the same amount of signal frequency content.

**3.2  Edge interpolation test**

To evaluate the practical advantages of different interpolation kernels, we used them to scale and rotate digital images. A simple 257x257 edge phantom $P$ was built as shown in Figure 4.

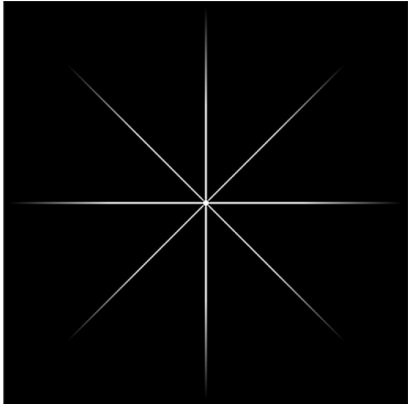

Figure 4: Edge phantom.

The phantom consisted of eight lines radiating from the center, with line intensity linearly decaying towards periphery. Our goal was to verify how edge strength and direction can be affected by interpolation under orthogonal image transforms. The transform were made out of zooming $Z$ and rotating $R$:

$$Z: z \to f * z, \quad f = 4/5$$
$$R: (x, y) \to \frac{1}{25}\begin{pmatrix} 24x & -7y \\ 7x & 24y \end{pmatrix} = \begin{pmatrix} x\cos(\alpha) & -y\sin(\alpha) \\ x\sin(\alpha) & y\cos(\alpha) \end{pmatrix}, \quad \alpha = \arcsin(7/25)$$
(Eq. 12)

The selection of rotation angle $\alpha = \arcsin(7/25)$ and zoom factor $f = 4/5$ was intentional to ensure that both $R$ and $Z$—as well as their inverses—would use finite-precision rational math, thus avoiding additional roundoff errors. The center of rotation was the phantom center. To cover the entire 360 degrees (roughly equal to $22\alpha$), and to keep the image within its original size, we applied repetitive zooms and rotations in the following manner:

$$P' = (Z * R * Z^{-1}R)^{11}(Z * R * Z^{-1}R)^{-11}P$$
(Eq. 13)

With this in mind, one would expect $P' = P$ if not for the interpolation errors. The visual representation of these errors in $P'$ is given in Figure 5.

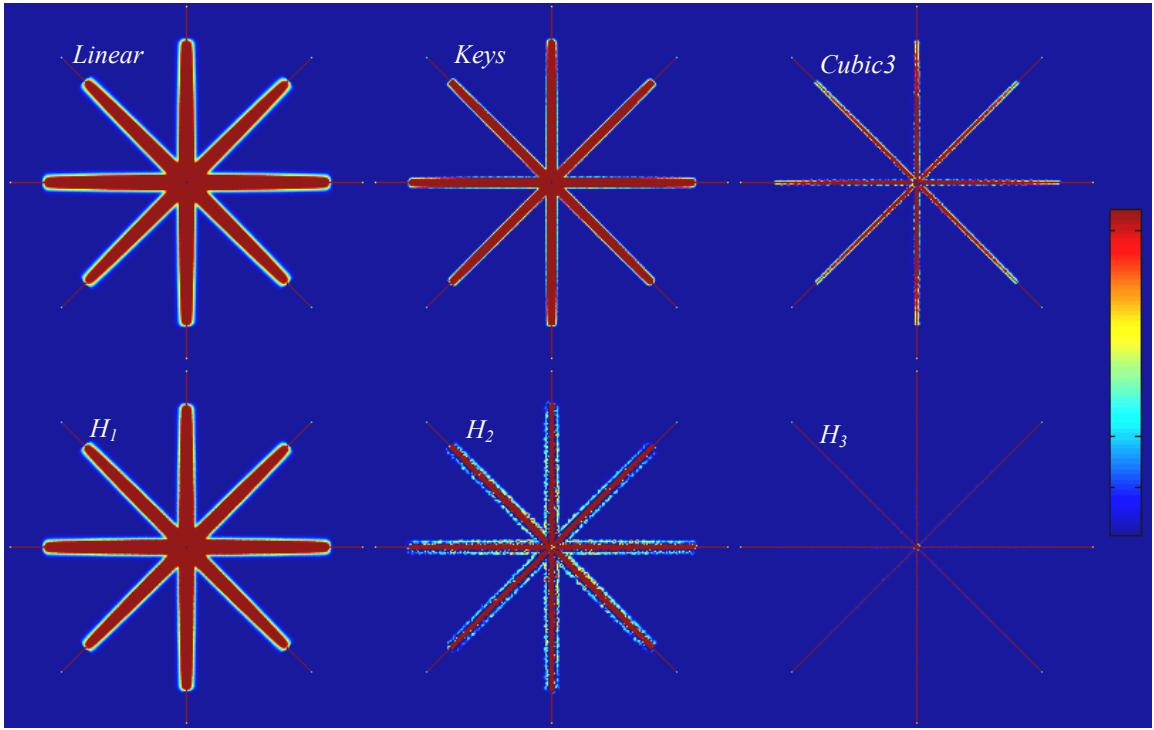

Figure 5: Visualizing interpolation errors for rotated and zoomed edge phantom $P'$. All six $P'$ images were artificially colored to show error magnitude, ranging from blue (minimal) to red (maximal). The errors in all images are shown on the same scale.

We conclude the following:

- Each column in Figure 5 corresponds to the same support size $L$ (1, 2, or 3), and as you can see, increasing $L$ decreases the interpolation errors (Figure 1, right).
- Optimal interpolation kernels $H_L$ (bottom row) for $L = 2$ and 3 perform visibly better than their popular counterparts (top row). This becomes obvious in case of $H_3$ ($L = 3$) when the errors virtually disappear. For $L = 2$, our optimal $H_2$ stands somewhere between the $h_{Keys}$ and $h_{Cubic3}$ kernels; the red error lines in $H_2$ become lighter (lower in magnitude) compared to $h_{Keys}$, but not yet as narrow as in $h_{Cubic3}$.

Note that our intentional choice of $R$ and $Z$ parameters—minimizing roundoff errors—was favoring popular $h_{Linear}$, $h_{Keys}$ and $h_{Cubic3}$ kernels, as using rational math in (Eq. 5). $L_2$-optimal kernels $H_L$, on the contrary, are based on irrational $Sinc(x)$ and therefore should have accumulated more errors. Yet their performance was clearly better.

## 3.3 Visual quality

To access interpolation quality visually, we tested $L_2$-optimal kernels with medical images known to have fine details and high contrast ratio. As we observed in our previous analysis, $L_2$-optimal kernel $H_2$ has the same degree of local frequency preservation for $L = 2$ as previously-known $h_{Cubic3}(x)$ on a larger $L = 3$ support. Therefore, we chose these two kernels and the previously known Keys $h_{Keys}(x)$ ($L = 2$) to validate our conclusions.

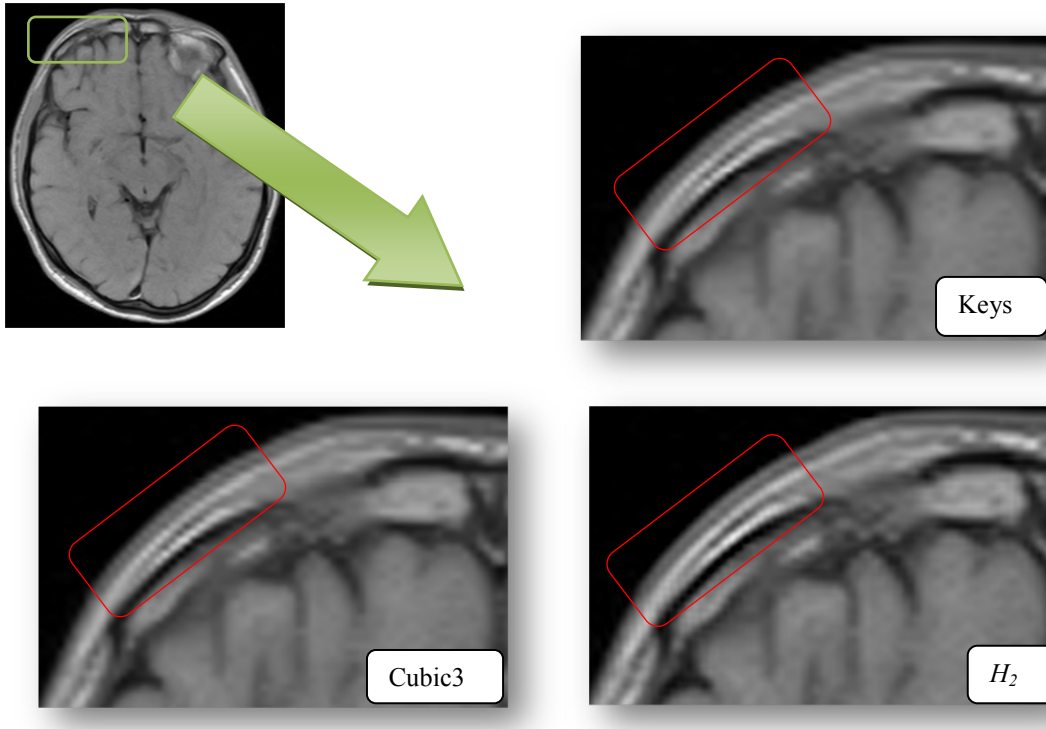

Figure 6: Zooming into MR image with different interpolation kernels: $H_2$ and Keys for support $L=2$, and Cubic3 for support size $L=3$. Both Keys and Cubic3 suffer from jagging artifacts on sharp image edges. $H_2$ is virtually artifact-free.

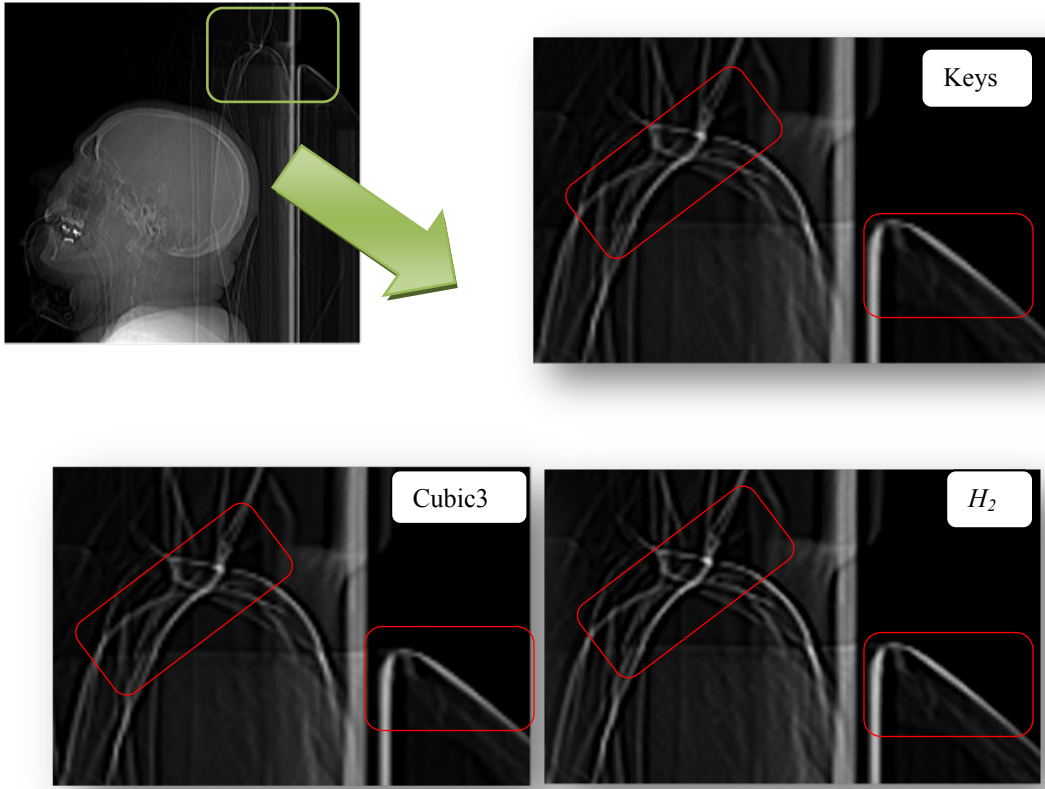

Figure 7: Zooming into MR image with different interpolation kernels: $H_2$ and Keys for support $L = 2$, and Cubic3 for support size $L = 3$. $H_2$ has visibly better quality with thin image details compared to Keys and Cubic3.

As a result, our visual comparisons confirmed higher interpolation quality delivered by $L_2$-optimal kernels.

## 4  Performance optimization

One true advantage of the previously used kernels (Eq. 5) is that they are simple to compute. Nonetheless, we would like to revisit the entire concept of computational simplicity from a different angle. It's important to realize that all signal-interpolating software relies on discrete sets of digital transforms. For instance, the transforms discreteness in digital imaging applications comes from two principal sources:

- Discrete zoom and rotation selections in the software interface. For example, even seemingly continuous mouse scrolls and wheel rotations, often associated with zooming, are implemented as discrete, incremental zooms with fixed zoom factors.
- More importantly, computers are discrete by definition and all computations are carried out with finite truncated precision.

In other words, one can easily challenge interpolation kernel continuity as hardly used or needed in signal-interpolation applications. Instead, kernels $h(x)$ should be implemented as discrete lookup tables (LUT) $g_K(x)$, storing kernel values rounded to a certain precision $K$:

$$g_K(x) = h([xK]/K) \qquad \text{(Eq. 14)}$$

where [] stands for nearest integer roundoff operator, and large integer $K$ defines the selected precision (lookup table size for a unit interval). In this case, the complexity of $h(x)$ formulas becomes completely irrelevant: all kernels for the same support size $L$ can be applied with the same constant time[5].

How can one choose LUT precision $K$ to guarantee sufficiently accurate signal interpolation? The following statements answer this question in the most general D-dimensional case.

**Definition 1:**

*Let $I(k_1,...,k_D)$ be a D-dimensional digital signal acquired with B bits per sample accuracy, $h(x)$ – the interpolation kernel, and K – the lookup table precision for tabulated kernel in (Eq. 14). Then we define permissible signal resolution $B_0$ as highest B such that, given $h(x)$ and K, signal $I(k_1,...,k_D)$ can be interpolated without errors by tabulated kernel $g_K(x)$ in (Eq. 14).*

**Theorem 2 (permissible signal resolution)**
*Let's assume that kernel $h(x)$ is differentiable and*

$$|h(x)| \le h < \infty$$
$$|h'(x)| \le \gamma < \infty, \quad 0 \le x \le L \qquad (Eq.\ 15)$$

*Then the following estimate for permissible signal resolution $B_0$ holds true:*

$$B_0 = -\log_2\left(2(L+1)^D h^D \left((1+\frac{\gamma}{hK})^D - 1\right)\right) \qquad (Eq.\ 16)$$

The proof for Theorem 2 is given in the appendix. Practically (Eq. 16) means that all signals with $B \le B_0$ will have exact interpolation with $K$-tabulated $h(x)$ from (Eq. 14). The size of kernel support $L$ and signal dimensionality $D$ are known beforehand. Kernel estimates $h$ and $\gamma$ can easily be found for any kernel—for instance, we can (Figure 2) safely assume $|h(x)| < h = 1.1$ for all kernels mentioned in this paper.[6] Therefore, lookup table precision $K$ is the only variable factor affecting $B_0$. We expect $K$ to be large, which brings us to the following asymptotic estimate:

$$B_0 = -\log_2\left(2(L+1)^D h^D \left((1+\frac{\gamma}{hK})^D - 1\right)\right) \xrightarrow[K \to \infty]{}$$
$$\xrightarrow[K \to \infty]{} -\log_2\left(2(L+1)^D h^D \frac{D\gamma}{hK}\right) = \log_2 K - \log_2\left(2(L+1)^D h^{D-1} D\gamma\right) \qquad (Eq.\ 17)$$

In other words, for large $K$ permissible signal resolution, $B_0$ behaves as a linear function of $\log_2 K$. To validate this result numerically, we made the plots of $B_0(K)$ for our six choices of kernels from (Eq. 5) and (Eq. 11), as shown in Figure 8.

---

[5] In the early computer age, when holding large LUTs in memory was too costly, recomputing interpolation kernels at each point was more practical.

[6] $L_2$-optimal kernels are not truly differentiable at integer points $x = n > 0$, but we can use their left and right derivatives there.

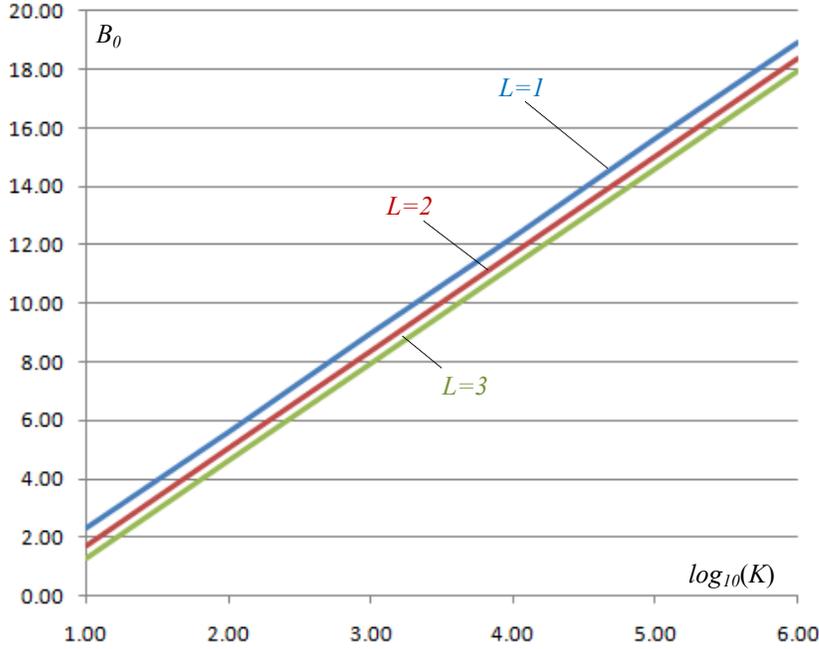

Figure 8: Functions $B_0(K)$ for six choices of $h(x)$ studied in this work. All $B_0(K)$ overlap for kernels with the same support size $L$. $log_{10}K = log_2K / log_210$, so the line slope is $log_210 = 3.3219...$

As you can see, linear $B_0$ plots in Figure 8 ideally conform to our asymptotical estimate in (Eq. 17). Note that higher support size $L$ results in lower $B_0$, as expected from (Eq. 17); this is the penalty for the higher interpolation quality. As we observed earlier from Figure 2, different interpolation kernels for the same $L$ may look very similar in the $x$ domain so that their $h$ and $\gamma$ estimates will be very close. This explains why different kernels of the same support size $L$ (such as $h_{Linear}(x)$ and $H_1(x)$) produced overlapping lines in Figure 7.

From a practical point of view, Figure 8 demonstrates that for $B = 8$ bit/sample signals LUT interpolation becomes error-free with $K$ as low as $10^3$. First of all, 8 bit/sample signals correspond to most popular conventional image and video formats (BMP, JPEG, MPEG), and their fast and accurate interpolation is one of the most demanded. Secondly, lookup table for $K = 1000$ and kernel support size $L = 3$ will have only $LK = 3000$ entries. If each entry is stored with 8-byte double precision, the total table size in memory will be about 192KB— a small amount that even smartphones can handle. However, at this minimal cost we can completely eliminate the issue of computational kernel complexity, applying any kernel as a lookup table.

More advanced applications such as medical imaging may require $B_0$ on the order of 12 or 14 bits/pixel. As Figure 8 demonstrates, this can be achieved with lookup tables of $K = 10^5$. For $L = 3$, this will require around 19.2MB of memory, which is still very affordable, taking less than an average CT or MR image series (order of 100 MB). Moreover, our estimate for $B_0$ in (Eq. 16) can often be improved to produce acceptable interpolations with smaller $K$:

- Some applications may tolerate less-than-exact interpolation accuracy, in which case the value of $K$ can be lowered (see the proof of Theorem 2).

- Dynamic signal range in local *L*-neighborhood (used for interpolation, (Eq. 3)) usually takes less than full *B* bits.
- Lookup table design in (Eq. 14) can be further enhanced at minimal computational cost. For instance, we can use linear interpolation between two nearest table values, to improve table accuracy and to reduce *K*.

To conclude, our kernel accuracy analysis proves that lookup tables can be used to implement interpolation kernels in the most performance-efficient way, independent on the $h(x)$ function complexity. Thus, the question of $L_2$-optimal kernel computational expense is entirely eliminated.

## 5 Conclusions

We introduced and derived a new family of digital signal interpolation kernels: $L_2$-optimal kernels $H_L(x)$, preserving most of the interpolated signal frequency content in $L_2$. This derivation was free of any non-essential constraints and produced the general equation for $H_L(x)$ on arbitrary support size *L* (Theorem 1). We also proved that $H_L(x)$ is continuous.

Our numerical and visual analysis confirmed that compared to the previously known results, $L_2$-optimal kernels deliver higher interpolation accuracy. In particular, we demonstrated that $L_2$-optimal kernel $H_2(x)$ on support $L = 2$ provides the same interpolation quality as previously known but suboptimal $h_{Cubic3}(x)$ on larger $L = 3$. The study of lookup table implementation proved that $L_2$-optimal kernels, just like any other, can be efficiently tabulated and computed regardless of the kernel function complexity.

Another interesting outcome of our work was the study of the frequency approximation error function $E(h)$, particularly with respect to the filter support size *L*. In $L_2$-sense, this function provides the exact lower bound on signal interpolation quality for any kernel $h(x)$. Our optimal kernels $H_L(x)$ correspond to the $E(h)$ minimum.

## 6 Appendix

### 6.1 Proof of Theorem 1

From the orthogonality of Fourier transform *F* (Parseval's theorem) and $F(Sinc(x)) = \Pi(t)$, taking into account finite support and symmetry of $h(x)$, we rewrite

$$E^2(h) = \int_{t=-\infty}^{+\infty}(F_h(t) - \Pi(t))^2 dt = \int_{x=-\infty}^{\infty}(h(x) - Sinc(x))^2 dx$$
$$= 2\int_0^L (h(x) - Sinc(x))^2 dx + 2\int_L^{+\infty} Sinc^2(x) dx = 2(E_1(h) + E_2)$$

(Eq. 18)

The second term $E_2$ does not depend on $h(x)$. The first term can be rewritten as

$$E_1(h) = \int_0^L (h(x) - Sinc(x))^2 dx = \sum_{k=0}^{2L-1}\left[\int_{k/2}^{(k+1)/2}(h(x) - Sinc(x))^2 dx\right]$$

(Eq. 19)

On each half-unit segment $[k/2, (k+1)/2]$ we change the integration variable as $x' = (-1)^k\left(x - \left\lfloor\dfrac{k+1}{2}\right\rfloor\right)$, to shift all integration segments to $[0, \frac{1}{2}]$:

$$E_1(h) = \sum_{k=0}^{2L-1}\left[\int_{k/2}^{(k+1)/2}(h(x) - Sinc(x))^2\,dx\right] =$$

$$= \sum_{k=0}^{2L-1}\left[\int_0^{1/2}\left(h\left(\left\lfloor\dfrac{k+1}{2}\right\rfloor + (-1)^k x\right) - Sinc\left(\left\lfloor\dfrac{k+1}{2}\right\rfloor + (-1)^k x\right)\right)^2 dx\right] \quad \text{(Eq. 20)}$$

$$= \int_0^{1/2}\left[\sum_{k=0}^{2L-1}\left(h\left(\left\lfloor\dfrac{k+1}{2}\right\rfloor + (-1)^k x\right) - Sinc\left(\left\lfloor\dfrac{k+1}{2}\right\rfloor + (-1)^k x\right)\right)^2\right] dx$$

At the same time from (Eq. 4), finite support, and symmetry of $h(x)$:

$$S(x) = \sum_{k=-L}^{L} h(x+k) = \sum_{k=0}^{2L-1} h\left(\left\lfloor\dfrac{k+1}{2}\right\rfloor + (-1)^k x\right) = 1, \quad 0 \le x \le 1 \quad \text{(Eq. 21)}$$

Therefore let's introduce functions

$$h_k(x) = \begin{cases} h\left(\left\lfloor\dfrac{k+1}{2}\right\rfloor + (-1)^k x\right), & x \in [k/2, (k+1)/2] \\ 0, & \text{otherwise} \end{cases} \quad \text{(Eq. 22)}$$

and

$$s_k(x) = Sinc\left(\left\lfloor\dfrac{k+1}{2}\right\rfloor + (-1)^k x\right) \quad \text{(Eq. 23)}$$

Then each $h_k(x)$ uniquely defines $h(x)$ on $[k/2, (k+1)/2]$, and

$$h(x) = \sum_{k=0}^{2L-1} h_k\left((-1)^k\left(x - \left\lfloor\dfrac{k+1}{2}\right\rfloor\right)\right)$$

From (Eq. 21) $\sum_{k=0}^{2L-1} h_k(x) = 1$, or $h_{2L-1}(x) = 1 - \sum_{k=0}^{2L-2} h_k(x)$, and substituting $h_{2L-1}(x)$ in (Eq. 20) yields:

$$E_1(h) = \int_0^{1/2}\left[\sum_{k=0}^{2L-2}(h_k(x) - s_k(x))^2 + \left(-1 + s_{2L-1}(x) + \sum_{k=0}^{2L-2} h_k(x)\right)^2\right] dx \quad \text{(Eq. 24)}$$

This is a well-defined variance optimization problem for a set of (2L-1) independent functions $h_k(x)$, k=0…2L-2. The minimum of $E_1(h)$ in (Eq. 24) should satisfy Euler's condition $\partial E_1 / \partial h_n = 0$ [9], leading to the following system of linear equations:

$$\sum_{k=0}^{2L-2} h_k(x) + h_n(x) = 1 + s_n(x) - s_{2L-1}(x), \quad n=0..2L-2 \tag{Eq. 25}$$

If we define $U(x) = \sum_{k=0}^{2L-2} h_k(x)$ and $W(x) = \sum_{k=0}^{2L-2} s_k(x)$, then by adding all the equations in (Eq. 25) we find

$$2LU(x) = (2L-1)(1 - s_{2L-1}(x)) + W(x), \text{ or}$$

$$U(x) = \frac{(2L-1)(1 - s_{2L-1}(x)) + W(x)}{2L}, \text{ and}$$

$$h_n(x) = 1 + s_n(x) - s_{2L-1}(x) - U(x) = 1 + s_n(x) - s_{2L-1}(x) - \frac{(2L-1)(1 - s_{2L-1}(x)) + W(x)}{2L} =$$
$$= \frac{1 - s_{2L-1}(x) + 2Ls_n(x) - W(x)}{2L} = s_n(x) + \frac{1}{2L}\left(1 - \sum_{k=0}^{2L-1} s_k(x)\right) \tag{Eq. 26}$$

Then $h_{2L-1}(x) = 1 - \sum_{k=0}^{2L-2} h_k(x) = s_{2L-1}(x) + \frac{1}{2L}\left(1 - \sum_{k=0}^{2L-1} s_k(x)\right)$, thus following the same equation (Eq. 26) for n=2L-1 From (Eq. 23)

$$h_n(x) = Sinc(x) + \frac{1}{2L}\left[1 - \sum_{k=0}^{2L-1} Sinc\left((-1)^{k+n} x + \left\lfloor \frac{k+1}{2} \right\rfloor + (-1)^{k+n+1} \left\lfloor \frac{n+1}{2} \right\rfloor\right)\right] = \tag{Eq. 27}$$
$$= Sinc(x) + T_n(x), \quad n/2 \leq x < (n+1)/2$$

This expression can be further simplified if we consider closely the argument of *Sinc* in $T_n(x)$:

$$a(k,n) = (-1)^{k+n} x + \left\lfloor \frac{k+1}{2} \right\rfloor + (-1)^{k+n+1} \left\lfloor \frac{n+1}{2} \right\rfloor$$

By simple substitution we verify the following for odd and even choices of k and n:

$$a(2p+1, 2m+1) = a(2p, 2m) = x + p - m = x + (k-n)/2 \tag{Eq. 28}$$
$$a(2p, 2m+1) = a(2p+1, 2m) = m + p + 1 - x = -x + (k+n+1)/2$$

In other words, even-numbered $a(k,n)$ in $T_{2m}(x)$ correspond to odd-numbered in $T_{2m+1}(x)$, and vice versa. Therefore

$$T_{2m}(x) = T_{2m+1}(x) = \frac{1}{2L}\left[1 - \sum_{k=0}^{2L-1} \operatorname{Sinc}\left((-1)^k x + \left\lfloor \frac{k+1}{2} \right\rfloor - (-1)^k m\right)\right]$$

And we finally write

$$H_L(x) = \sum_{m=0}^{L-1} h_m(x) = \begin{cases} \operatorname{Sinc}(x) + \frac{1}{2L}\left[1 - \sum_{k=0}^{2L-1} \operatorname{Sinc}\left((-1)^k x + \left\lfloor \frac{k+1}{2} \right\rfloor - (-1)^k m\right)\right] \\ m \leq x < m+1, \quad m = 0...L-1 \end{cases} \quad \text{(Eq. 29)}$$

It is interesting to observe that $H_L(x)$ is continuous, although we never required any continuity in our derivation. Indeed, consider $a(k,n)$ in (Eq. 28): they will equal to $x$ only when $p = m$, and they will be $-x$ only when $m + p + 1 = L$. This means that for any integer $x = i$ function $T_n(x)$ will contain one positive $\operatorname{Sinc}(0) = 1$ and one negative $-\operatorname{Sinc}(0) = -1$ term, cancelling each other and the remaining $\operatorname{Sinc}(l) = 0$ for some integer $l \neq 0$. Therefore, for any integer $i$ and any $m > 0$, $T_m(i) = 0$, and piecewise functions in (Eq. 29) continuously connect to each other at the ends of their intervals.

∎

## 6.2  Proof of Theorem 2

**Lemma 1**

*Consider two sets of variables $e_i$ and $h_i$, $i = 1..D$, limited in absolute values by some constants $e$ and $h$:*

$$|h_i| \leq h, \quad |e_i| \leq e \quad \text{(Eq. 30)}$$

*Then the following inequality holds true*

$$\left|(h_1 + e_1)(h_2 + e_2)...(h_D + e_D) - h_1 h_2 ... h_D\right| \leq (h+e)^D - h^D \quad \text{(Eq. 31)}$$

**Proof**

Expanding the terms, we write (using $\binom{D}{n}$ for binomial coefficient):

$$\left|(h_1 + e_1)(h_2 + e_2)...(h_D + e_D) - h_1 h_2 ... h_D\right| = \left|\sum_{\substack{1 \leq i_1,...,i_n \leq D \\ 1 \leq j_1,...,j_{D-n} \leq D \\ 1 \leq n \leq D}} e_{i_1}...e_{i_n} h_{j_1}...h_{j_{D-n}}\right| \leq \quad \text{(Eq. 32)}$$

$$\leq \sum_{\substack{1 \leq i_1,...,i_n \leq D \\ 1 \leq j_1,...,j_{D-n} \leq D \\ 1 \leq n \leq D}} |e_{i_1}|...|e_{i_n}||h_{j_1}|...|h_{j_{D-n}}| \leq \sum_{n=1}^{D} \binom{D}{n} e^n h^{D-n} = (h+e)^D - h^D$$

In particular, (Eq. 31) becomes exact equality when $|h_i| = h$ and $|e_i| = e$.

∎

**Lemma 2 (distorted interpolation)**

Let $I(k_1,...,k_D)$ be an arbitrary D-dimensional digital signal, and $J(k_1,...,k_D)$ its D-dimensional interpolation, computed with interpolation kernel $h(x)$ from D-dimensional version of (Eq. 3):

$$J(x_1,...,x_D) = \sum_{k_1=0}^{L}...\sum_{k_D=0}^{L} I(k_1,...,k_D) h(x_1-k_1)..h(x_D-k_D)$$

Let $|I(k_1,...,k_D)|<M$ and $|h(x)|<h$. Then, if we distort kernel $h(x)$ with some error $e(x)$ as $h'(x)=h(x)+e(x)$, $|e(x)|<e$, the error in the distorted interpolation $J'(x_1,...,x_D)$ obtained with $h'(x)$ can be estimated as follows

$$|J(x,y) - J'(x,y)| < M(L+1)^D \left( (h+e)^D - h^D \right) \qquad \text{(Eq. 33)}$$

**Proof**

We can prove Lemma 2, applying Lemma 1 to estimate the interpolation error when using distorted kernel values. Let us consider the original interpolation kernel $h(x)$, and the distorted $h'(x)=h(x)+e(x)$. Then we define the original and distorted interpolation results as:

$$J(x_1,...,x_D) = \sum_{k_1=0}^{L}...\sum_{k_D=0}^{L} I(k_1,...,k_D) h(x_1-k_1)..h(x_D-k_D)$$

$$J'(x_1,...,x_D) = \sum_{k_1=0}^{L}...\sum_{k_D=0}^{L} I(k_1,...,k_D) h'(x_1-k_1)..h'(x_D-k_D) =$$

$$= \sum_{k_1=0}^{L}...\sum_{k_D=0}^{L} I(k_1,...,k_D) [h(x_1-k_1)+e(x_1-k_1)]..[h(x_D-k_D)+e(x_D-k_D)]$$

Therefore, using $|I(k_1,...,k_D)|<M$ and Lemma 1, we estimate:

$$|J(x,y) - J'(x,y)| \leq$$

$$\leq M \sum_{k_1=0}^{L}...\sum_{k_D=0}^{L} \left| [h(x_1-k_1)+e(x_1-k_1)]..[h(x_D-k_D)+e(x_D-k_D)] - h(x_1-k_1)...h(x_D-k_D) \right| \leq$$

$$\leq M(L+1)^D \left( (h+e)^D - h^D \right)$$

∎

Now we can prove Theorem 2. For distorted interpolation to be exact, it is enough to require $|J(x,y)-J'(x,y)|<\frac{1}{2}$, so that after the nearest-integer roundoff $J'(x,y)$ would become identical to $J(x,y)$. To achieve this, it is sufficient to require $M(L+1)^D \left( (h+e)^D - h^D \right) \leq \frac{1}{2}$ or

$M \leq \frac{1}{2(L+1)^D \left( (h+e)^D - h^D \right)}$. If signal samples are acquired with $B$ bits per sample resolution, then $M = 2^B$ and

$$B \leq -\log_2\left(2(L+1)^D\left((h+e)^D - h^D\right)\right) = -\log_2\left(2(L+1)^D h^D\left((1+\frac{e}{h})^D - 1\right)\right) \quad \text{(Eq. 34)}$$

In this estimate, the only unknown parameter *e* depends on the kernel distortion, which in our case is a function of the lookup table precision *K*:

$$e = e(K) = \max_{0 \leq x \leq L} |g_K(x) - h(x)| = \max_{0 \leq x \leq L} |h([xK]/K) - h(x)| \quad \text{(Eq. 35)}$$

But for finite-differentiable *h(x)* we can estimate, with some *t* in the ([xK]/K,x) interval:

$$|h([xK]/K) - h(x)| = |h'(t)| \, |[xK]/K - x| \leq \gamma/K \quad \text{(Eq. 36)}$$

Substituting this estimate for *e* in (Eq. 34), we obtain

$$B \leq -\log_2\left(2(L+1)^D h^D\left((1+\frac{\gamma}{hK})^D - 1\right)\right) \quad \text{(Eq. 37)}$$

which proves Theorem 2.

∎